# Online Learnability of Statistical Relational Learning in Anomaly Detection

Magnus Jändel, Pontus Svenson and Niclas Wadströmer
Swedish Defence Research Agency
Sweden
Contact: magnus.jaendel@foi.se

*Abstract*— Statistical Relational Learning (SRL) methods for anomaly detection are introduced via a security-related application. Operational requirements for online learning stability are outlined and compared to mathematical definitions as applied to the learning process of a representative SRL method – Bayesian Logic Programs (BLP). Since a formal proof of online stability appears to be impossible, tentative common sense requirements are formulated and tested by theoretical and experimental analysis of a simple and analytically tractable BLP model. It is found that learning algorithms in initial stages of online learning can lock on unstable false predictors that nevertheless comply with our tentative stability requirements and thus masquerade as bona fide solutions. The very expressiveness of SRL seems to cause significant stability issues in settings with many variables and scarce data. We conclude that reliable anomaly detection with SRL-methods requires monitoring by an overarching framework that may involve a comprehensive context knowledge base or human supervision.

I. INTRODUCTION

*A. Application scenario*

Real-time abnormal behavior detection is an important component of many security applications today. A simple example is a facility surveillance system, where one wants to detect anomalies and take appropriate action to either prevent them or mitigate their consequences. Generally, it will be necessary to combine information from many different sources in order to detect the anomalies; hence fusion is a necessary component of the surveillance systems. The anomalies of interest could range from terrorists planning an attack to more pedestrian situations such as theft by employees or customers in a port facility. Even though it is the terrorist cases that presently get the most attention, it is important not to forget the possibilities to detect everyday crime using the same systems.

The research described in this paper is motivated by applications in EU FP7 projects on supply chain security [1] [2] [3].There are several different aspects of supply chain security [4] [5]. Perhaps most obvious is the scanning and screening of cargo and bags that take place in airports. Similar scanning is done on containers in ports and on postal items in sorting facilities. Ideally all cargo/containers should be scanned but economic realities favor a risk-based approach, where a risk index is computed for each cargo item, and only those with a high risk index are subjected to the most expensive scanning.

One way of computing risk indicators is to use abnormal behavior detection on all available data regarding the item. This can include information from sensors and other sources. For instance, relevant information for containers might include senders and receivers of all goods packaged inside the container, the owner and crew of the ship transporting it, and the journey details of the ship. This set of information is highly heterogeneous, which complicates the task of fusing and analyzing it. Furthermore, the analysis of what constitutes a threat and not is not static, but will change dynamically. Hence, there is a need for methods for fusion of heterogeneous information and anomaly detection that can adapt to changing threat profiles.

An equally important application of anomaly detection for supply chain security is in facility protection and surveillance. While abnormal behavior detection in crowds is more common in mass transport security applications, several similar issues arise also in facility surveillance. Consider a port where both passengers and employees are at large. Anomaly detection can be used to detect individuals who appear to be preparing criminal acts. Here, too, the threat patterns are evolving quickly and there is a need for online learning to adapt the anomaly detection system to the changing conditions.

A full description of an anomaly detection system is out of the scope of this paper. We will however assume that the system consists of a set of sensors that give information about the current state of the world. This information is fused to produce higher-level situation descriptions which are fed into the anomaly detection system. In this paper we consider anomaly detection to be a form of classification where the system learns from a training set including both normal and anomalous situations. The trained system classifies situations as either normal or anomalous and operators are alerted to detected anomalies. Note that the availability of labeled data including both normal and anomalous events often is an issue in the supervised classification approach to anomaly detection employed here. For simplicity we assume that there is only one type of anomaly although practical applications typically include many anomaly types. Alternative paradigms, not considered here, include unsupervised learning where low-probability events are considered anomalous.

The core of the anomaly detection system is assumed to be based on Statistical Relational Learning (SRL) [6]. A key requirement is that the classification performance of the continuously learning anomaly detection system is stable. In this paper we therefore investigate the online stability of SRL in the context of our application scenario.

*B. Statistical Relational Learning*

Extending machine learning to take account of the rich relational structure of real-world problems is the goal of the emerging science of SRL (see [6] for a comprehensive overview). Many different methods have been proposed including Bayesian Logic Programs (BLP) [6] which is the SRL method selected for the present investigation. SRL methods describe a very broad class of probability density functions (PDFs) using a two-tier representation consisting of a structural part and a parametric part. The structural part is usually written in the language of first order logic while the parametric part expresses the statistical significance of the relational structure. An explicit representation of a PDF, typically in the form of a Bayesian network or a Markov random field, is generated from the SRL models in response to a query. This explicit PDF representation is then used to compute the answer to the query, e.g., the probability that an observed event is anomalous. SRL enables potentially a wide range of powerful applications but the profuse expressiveness of SRL is also the source of the stability issues that are discussed in the following.

*C. Bayesian Logic Programs*

A BLP consists of a set of ground terms, a set of logical atoms and a set of so-called Bayesian clauses that represent relations. The ground terms represent known objects. Logical atoms are well-founded facts and relations including, in particular, universally quantified relations. The ground terms together with the logical atoms form a base-line logical knowledge base $KB_0$. A Bayesian clause describes a dependency relation in the form of a logical clause and is associated with probabilistic parameters describing the statistical significance of the relation. The statistical parameters take the form of a conditional probability table (CPT). The set of Bayesian clauses constitute an extended knowledge base $KB_B$. A total knowledge base $KB$ is formed by merging $KB_0$ and $KB_B$.

Consider a new object e.g. $Q_{23}$ that has been discovered by the surveillance system. First the new ground term and the observed facts are added to the knowledge base. Then the BLP system estimates the degree of anomalousness of the object in three steps.

    A) The knowledge base $KB$ is first used to construct a logical proof of the predicate Anomalous($Q_{23}$). For simplicity we assume that precisely one such proof is found. Ref. [6] describes how combination rules are used for handling multiple proofs. The logical proof is expressed as a set of Horn clauses [1]. In the following we will use the implication format of a Horn clause and it is understood that "clause" means "Horn clause".

    B) The set of clauses that constitute the proof can be viewed as a network with predicates as nodes and inference relations as edges. By transforming predicates to random variables and inference relations to statistical dependency relations, the proof is converted to a Bayesian network that is called the *support network* of the query. Each node in the support network inherits a CPT from the Bayesian clause that is the template of the node. The CPT codifies the statistical dependency between the dependent variable and its parents. The specific statistical parameters of the support network are hence instantiated from the generic parameters of the BLP.

    C) The probability of $Q_{23}$ being anomalous is finally estimated by inference in the support network.

In summary, queries to the BLP system are requests for estimates of the marginal probability distribution of random variables. The system interprets random variables as predicates, attempts to prove the predicates, converts the proof to a support network and estimates the requested marginal probability distribution from the support network.

The main challenge is how to learn Bayesian clauses from data. Presently this is done in a two-stage loop where the space of relations first is explored for the purpose of identifying candidate sets of clauses. The second step strives to find optimal statistical parameters for each candidate set of clauses. By selecting the best set of Bayesian clauses as the starting point of the next iteration, the learning process eventually converges. BLP learning is further discussed in section II A.

*D. Introduction to online learning in SRL*

SRL learning processes including the BLP learning algorithm that is outlined in the previous section are primarily designed for batch processing where the learning process is executed once using all available data. In many applications, including our surveillance scenario, it is more natural to consider online learning where the system learns continuously from a stream of incoming data while it simultaneously must respond to queries. Disregarding computational efficiency and memory issues it is possible to run a batch learning process on all the accumulated data as soon as new data arrives. It is, however, vital that online learning is stable in security-critical systems. A single new training example could otherwise cause a sudden performance degradation. Definitions of stability are discussed from an operational point of view in section I E and in the context of learning theory in section II B.

---

[1] Recall that a Horn clause is a disjunction of literals, where all except at most one literal is negated. It can also be viewed as an implication, where the literal that was negated is implied from the conjunction of the others.

Discussion of SRL online learning in the literature seems to focus on computational complexity aspects. Online learning of statistical parameters in Markov Logic Networks (MLN) is considered in [7], [8] and [9]. Huynh and Mooney address online learning of both structural and statistical aspects of MLN and introduce a regularization method that suppresses the creation of new clauses while learning online [10].

### E. Operative requirements for online stability in anomaly detection

Generic critical properties of anomaly detection systems are the false alarm rate $P_{false}$ and the missed alarm rate $P_{missed}$. The former quantity is the probability of erroneously detecting an anomaly in a normal situation while the latter quantity is the probability of not detecting an anomaly in an anomalous situation. $P_{false}$ is the same as False Positive Rate in machine learning, while $P_{missed}$ is the False Negative Rate. The anomaly detection system is useful only if $P_{false}$ and $P_{missed}$ fall below thresholds $T_{false}$ and $T_{missed}$ respectively where threshold values depend on contingent operative requirements.

Furthermore, the system should be able to run with reliable good performance between maintenance check-points. We assume here that the system runs unattended except for regularly scheduled maintenance occasions when an operator monitors the performance of the system and decides if it can be relied on. This means that the online learning algorithm should comply with performance thresholds for at least $m^*$ new training examples where $m^*$ is given by the maximum rate of incoming training examples and the maintenance period.

Assuming that $m$ training examples have been processed initially practical requirements for online stability hence are that,

$$\forall m': 0 < m' \leq m^* \quad P_{false}(\mathbf{S}_{m+m'}) < T_{false}, \quad (1)$$

and

$$\forall m': 0 < m' \leq m^* \quad P_{missed}(\mathbf{S}_{m+m'}) < T_{missed}, \quad (2)$$

where $\mathbf{S}_m$ means a training set consisting of $m$ samples.

An anomaly detection system such that,

$$\forall m : P_{false}(\mathbf{S}_{m+1}) \leq P_{false}(\mathbf{S}_m), \quad (3)$$

and

$$\forall m : P_{missed}(\mathbf{S}_{m+1}) \leq P_{missed}(\mathbf{S}_m), \quad (4)$$

is guaranteed to be stable for any value of the operational parameters $T_{false}$, $T_{missed}$ and $m^*$. A system with this property is *universally stable* since adding a new training example to a system with satisfactory performance will not, under any circumstances, impair the performance of the system. Universally stable systems can be trusted to learn online indefinitely.

For reasons that are spelled out in the following, universal stability in SRL online learning is a chimera at the present state of the art. We can strive for enhancing online stability but never reach the epitome of universal stability.

## II. ANALYSIS OF ONLINE STABILITY

This section provides a more detailed description of BLP learning, a brief review of online stability in machine learning and discussion of a simplified BLP model where sources of online instability are identified and discussed.

### A. BLP Learning

In the following we review BLP learning as described in [6] and [11]. The goal of BLP learning is to infer a BLP from a set of training examples $\mathbf{S}$. Each training example $S_i \in \mathbf{S}$ consists of a list of random variables where values may be assigned to some of the variables e.g. $S_{56}$ = {flies(*Donald*) = ?, colour(*Donald*) = *white*, father(*Quackmore*, *Donald*) = *true*, mother(*Hortense*, *Donald*) = *true* }.

Inductive Logic Programming (ILP) [12] is first applied to create the initial set of clauses from the data. For this purpose each training example is viewed as list of logical assertions where random variables are transformed to predicates and the assigned values are ignored. The training example $S_{56}$ becomes for example the assertion {flies(*Donald*), colour(*Donald*), father(*Quackmore*, *Donald*), mother(*Hortense*, *Donald*)}. Using the transformed data set and optionally background knowledge in $KB_0$, the ILP engine generates a set of universally quantified Horn clauses codifying generic relations that can be learnt from $\mathbf{S}$ and $KB_0$. Note this set of clauses typically is not unique.

ILP delivers hence the structure of an initial BLP where each Horn clause becomes a Bayesian clause with yet unknown statistical parameters. Bayesian support networks are generated by instantiating the universally quantified variables with ground terms from the training set. The statistical parameters of the Bayesian clauses are computed by maximizing the combined likelihood of the support networks with respect to $\mathbf{S}$ (now taking variable values into account). The value of the combined likelihood for optimal statistical parameters becomes the utility of the initial BLP.

Refinement operators are now applied to the initial BLP so that a set of candidate BLPs are generated. In the presently studied implementation of BLP [11] two refinement operators are employed that, applied to the body of a clause, either adds or removes a universally quantified predicate. Viewed as a set of Horn clauses, each candidate is required to be consistent with $KB_0$. Viewed as a BLP, each candidate is scored by computing statistical parameters as before and evaluating utility by summing up the combined maximum likelihood of the support networks. The candidate with the highest utility is selected as the next preferred solution.

The learning process consists hence of a series of iterations where refinement operators are applied and the best candidate

BLP is selected. When no further significant improvement in utility is achieved, the learning process is deemed to have converged. Note that this process must be constrained so that only acyclic support networks are generated and that the maximum likelihood computation must handle combining rules. This section is only intended as a brief overview of BLP learning and rests completely on Ref. [6] and [11] that should be consulted for further details.

*B. Formal stability conditions for online learning*

Ross and Bagnell [13] provide an up-to-date review on what is known about stability conditions for online learnability in the context of the General Setting of Learning (GSL) [14]. In the online version of GSL an algorithm **A** observes a sequence of data points $z_1, z_2...z_m \in Z$. Each time the algorithm has received a subsequence $z_1, z_2...z_{i-1}$ it must select a hypothesis $h_i \in H$ and suffer a penalty given by the loss function $f: H \times Z \to R$ applied to the next data point $z_i$. Symbols $Z$ and $H$ denote the space of data and hypotheses respectively.

The problem of online stability in anomaly detection, as discussed in section I E, is strictly not a GSL problem since there are several different objectives and the problem therefore belongs to the class of multiobjective optimization problems [15]. To transform our stability requirements to a GSL problem we must define a computationally feasible loss function that captures the combined operational constraints expressed in terms of available data.

Ross and Bagnell define the regret of the sequence of hypotheses in GSL as,

$$R_m = \sum_{i=1}^{m} f(h_i, z_i) - \min_{h \in H} \sum_{i=1}^{m} f(h, z_i). \quad (5)$$

An algorithm has no regret at convergence rate $\varepsilon_{regret}(m)$ if for any sequence of data and for all m,

$$R_m / m \leq \varepsilon_{regret}(m), \quad (6)$$

where the convergence rate is a monotonously non-increasing function of $m$ and $\varepsilon_{regret} \to 0$ as $m \to \infty$. If an algorithm has no regret the problem is online learnable.

Ross and Bagnell discuss online learnability for the major classes of machine learning algorithms. It is not known if any SRL method is online learnable according to the definition of Ross and Bagnell. Even if we assume that an SRL algorithm has no regret and hence is online learnable we would only be able to say that the system will comply with operational stability requirements for sufficiently large $m$ without knowing how much data that actually is needed for a given system and situation. To assess if the system is online stable for any specific finite $m$ we also need an estimate of $\varepsilon_{regret}(m)$. The convergence rate is, however, known only for some special classes of algorithms notably those with a convex loss function. Convexity means that there are no local optima so that a greedy gradient ascent algorithm is guaranteed to find the global optimum. SRL problems typically feature a jagged fitness landscape that violates the convexity condition. The BLP learning process that is outlined in the previous section involves greedy gradient search with respect to both structure and parameters. Since it is unknown if BLP is online learnable and the convergence rate anyway is unknown it appears that it presently is difficult to use the formal online stability conditions of Ross and Bagnell as a basis for testing the practical online stability of BLP applications.

*C. A tentative rule of thumb for SRL online stability*

Because of the great potential advantages of using SRL methods it is tempting to ignore the lack of formal stability assurance and embrace a more cavalier approach according to the following sketch. Since real-valued statistical parameters always will vary under the impact of accumulating training samples it appears that the permanence of the "digital" clause structure is the best signature of stability. The clause structure should hence be monitored according to the following guidelines,

*1) Run the online learning process until the structure of the solution is stable for a time that is significantly longer than the maintenance interval.*

*2) Make sure that only minor refinements of the structure have occurred over the last few modifications of the solution. It is reasonable to assume that only small simpifications or specializations of the logical structure occurr as the SRL system converges to a bona fide solution.*

*3) Monitor the life time (the number of new online training samples that a structure survives). Certify that the life time of consecutive structures increases since this can be assumed to be a signature of approaching a valid solution.*

Our tentative rule of thumb amounts to accepting the solution for operational use if all of the conditions 1-3 are satisfied. Note that the requirement 2) of small structural changes only is used for verifying stability. Early learning phases may include large structural changes where the (unstable) system breaks out of local minima.

III. ANALYSIS OF ONLINE STABILTY IN A SIMPLIFIED MODEL

For the purpose of better understanding online stability in BLP and similar SRL concepts we will discuss a highly simplified model where stability issues are more tractable than in generic SRL learning. This toy model will also be the basis of numerical experiments where we test the rule of thumb in the previous section.

*A. A toy model*

Consider a set of n+2 random variables $X_A, X_0, X_1, ... X_n$ where each variable has a binary domain $X_i \in \{0,1\}$. The ground truth PDF is given by the following description. All variables with the exception of $X_A$ are independent with a uniform probability distribution. The anomalousness variable $X_A$ depends on $X_0$ so that $X_A$ with 80% probability has the same value as $X_0$.

A training set $\mathbf{S}_m$ consists of $m$ complete samples of all variables drawn from the ground truth PDF. We are interested in learning a BLP from the training set and then predict the value of $X_A$ from the values of the other variables in a test sample.

To prepare for BLP learning we consider each training example as a set of values $\{x_A(y_i), x_0(y_i), x_1(y_i), \ldots x_n(y_i)\}$ where $x_j(y_i)$ is the value of random variable $X_j(y_i)$, $y_i$ is the value of a dummy random variable $Y$ and $i$ is the index of the example. The Bayesian clause that fully models the ground truth probability distribution of relevance for the anomalousness indicator is,

$$X_A(Y) := X_0(Y), \qquad (7)$$

where $Y$, in the structural perspective, is a universally quantified logical variable representing a training example label and a CPT describes the relevant probability distribution. In addition to (7) there are $n$ redundant independent binary variables. This will in the following be called $\mathbf{BLP_T}$, the ground truth BLP.

Consider now a BLP learning process according to section II A where the initial model $\mathbf{BLP_0}$, for good measure, is taken to be identical to $\mathbf{BLP_T}$. Note that the learning process outlined in section II A normally would generate multiple initial clauses. For the purpose of creating an analytically tractable problem we assume here that the initial structure learning algorithm has found the ideal starting point and that hence just one clause is used in the learning process.

A sufficiently small initial training set includes redundant variables that accidentally are strongly correlated with $X_A$ and the total likelihood of the training set is improved by applying the expanding refinement operator to include such variables in the body of the clause as for example in the following sequence,

**BLP1**: $X_A(Y) := X_0(Y), X_{17}(Y)$

**BLP2**: $X_A(Y) := X_0(Y), X_{17}(Y), X_{47}(Y)$

….

**BLP5**: $X_A(Y) := X_0(Y), X_{17}(Y), X_{34}(Y), X_{47}(Y), X_{61}(Y)$

Each step increases the combined likelihood of the training set and is therefore a valid learning iteration according to section II A. In the present batch of training examples, a set of redundant variables in the body of **BLP5** $\{X_{17}, X_{34}, X_{47}, X_{61}\}$ just happens to code for the values of $X_A$ in the training set with 100 % accuracy since each combination of values for the set either maps to a given value of $X_A$ or does not occur in the training set. Applying the reductive refinement operator in the next learning iteration hence produces the 100% accurate BLP,

**BLP6**: $X_A(Y) := X_{17}(Y), X_{34}(Y), X_{47}(Y), X_{61}(Y)$.

Applying **BLP6** to a test set obviously gives much worse performance than $\mathbf{BLP_T}$ thus illustrating that BLP learning, at least in some situations, might reduce performance by overfitting to statistical fluctuations in the training sample.

### B. False predictors

The set $\{X_{17}, X_{34}, X_{47}, X_{61}\}$ together with the associated CPT is an example of a false predictor where a set of variables accidentally codes for another variable in a training set although the variables really are independent. In this section we will analyze how false predictors impact the BLP online learning process.

To facilitate this analysis we shall further simplify the toy model by assuming that the CPTs that are associated with each Bayesian clause consist only of binary-valued conditional probabilities $p \in \{0,1\}$. The *structure* of the single Bayesian clause of our toy model is represented by a list of body variables such as $\{X_{17}, X_{34}, X_{47}, X_{61}\}$. We further define a *pattern* as a bit pattern of body variable values e.g. $\{X_{17}=1, X_{34}=0, X_{47}=1, X_{61}=0\}$. The binary CPT of the Bayesian clause with $s$ variables in the body is hence represented by $2^s$ rows each row consisting of a pattern and an associated conditional probability $p \in \{0,1\}$ where p=1 means that $X_A=1$ with certainty. The total number of different CPTs for a given structure of size $s$ is countable in this model and amounts to $2^{2^s}$.

The variables $X_1, \ldots X_n$ that are decoupled from $X_A$ are called *redundant variables*. A *redundant structure* is a structure that consists of redundant variables. A false predictor is a Bayesian clause consisting of a redundant structure and a CPT where the clause classifies the values of $X_A$ in the training set with 100 % accuracy. Under the assumptions of the toy model, the average number of false predictors containing $s$ variables drawn from a set of $n$ variables $\{X_1, \ldots X_n\}$ and applied to a training set consisting of $m$ examples is,

$$N_{nsm} = \binom{n}{s} 2^{2^s - m}. \qquad (8)$$

The binomial coefficient in the equation corresponds to the number of ways that a structure of $s$ variables can be selected from a total of $n$ variables. The $2^{2^s}$ factor corresponds to the total number of CPTs for a given structure. Each new training example has an even chance of falsifying each of the false predictors thus explaining the negative factor $m$ in the exponent. By approximating the binomial coefficient, it is easy to show that for large $n$ and under certain other conditions $N_{nsm} \sim 2^{n+2^s - m}$. False predictors are hence ubiquitous e.g. for $n = 20$, $s = 10$ and $m = 1000$. Note that the $2^s$ term in the exponent of (8) is characteristic for the rich hypothesis space of relational algorithms and would be absent for non-relational machine learning methods.

If a false predictor has been established as the preferred hypothesis it will be challenged by further online learning. Each new training example will, under the assumptions of the toy model, have a 50 % chance of falsifying the false predictor. A false predictor survives hence in average for two online learning cycles under the assumptions of the toy model.

During initial online learning there will be a phase where the fitness landscape of the loss function includes many false minima each associated with a false predictor. This phase will be called *the false predictor phase*. As the online learning process proceeds, the population of false predictors is decimated. The life time of each selected hypothesis is short and the learning process will rapidly switch between false optima. Relevant solutions are found only after almost all false predictors have been falsified. In the toy model there is a 50 % probability that all false predictors are falsified after approximately (to leading order in *n*),

$$m_{1/2} = 2^{(\frac{n}{2}+1)}, \qquad (9)$$

training examples. In generic SRL problems it will in general be impossible to estimate how much data that is required to suppress false predictors since prior knowledge about the true joint probability distribution typically not is available.

From the toy model we shall endeavor to extrapolate some generic insights about SRL online learning. In particular, we evaluate the rule of thumb in section II C by studying the evolution of structure in online learning. In preparation for this investigation quantities that characterize the structure life cycle are defined.

The *life time* of a structure is the number of training examples in online learning from the point where the structure is selected by the learning algorithm to the point where it is discarded and replaced by a different structure. The minimum life time is one. A redundant structure will be more long-lived than the individual Bayesian clauses that incorporate the structure since each structure correspond to many Bayesian clauses with different CPTs each having some probability of being a false predictor.

The *structural size* is a measure of the logical complexity of a BLP. In our toy example the structural size is simply the number of variables in the body of the clause.

The *structural distance* between two BLP models is the minimum number of refinement operations that is required to transform one of the models to be identical to the other model. For two Bayesian clauses with the same head, the structural distance is just the number of predicates that differ between the clauses.

The *hop size* is the difference in structural distance between two consecutive solutions.

As the number of training examples increases during online learning, false predictors will be depleted at the same rate independently of the structure but smaller structures are associated with fewer CPTs and will hence run out of options sooner. Starting from a structurally simple hypothesis, we expect the average structural size of the preferred solution to increase gradually during online learning in the false predictor phase. In our example, the structural size would suddenly collapse when the true solution is found. In general we expect a sharp shift in structural size as the online learning process exits from the false predictor phase and transits from a highly specialized false predictor to a relevant solution.

The life time of the selected redundant structure will also increase as the training set expands. Larger structures correspond to a larger supply of possible CPTs that will require more training examples to falsify. Increased structural life time is therefore not per se an indicator of convergence to a stable and relevant solution.

We expect the average hop size to be fairly constant while the learning process is in the false predictor phase. The learning process favors minimal hops and the distribution of false predictors over the fitness landscape should be uniform.

Let us now relax the toy model restriction to binary CPTs and allow real-valued probabilistic parameters. We expect the same general behavior but with a smoother fitness landscape since the expanded model is less constrained. One single training example will now not be able to falsify a false predictor but will only weaken the likelihood of the hypothesis thus extending the life span. A more complex relational structure should not fundamentally change the picture. The size and life time of preferred redundant structures should still increase with increasing size of the training set. Hence we expect that the behavior of false predictors that we discussed in the context of the toy model qualitatively will remain the same in generic online learning of BLP and other similar SRL models.

IV. EXPERIMENTAL SETUP AND RESULTS

To confirm the theoretical arguments in section III we have performed numerical experiments where the toy model described in section III A is used for generating training samples. We apply a BLP learning algorithm as described in section II A modified with the constraint to binary CPTs. The loss function is the inverse of the likelihood of the training data according to the hypothesis. For each training set a best fitting hypothesis is found. The training set is incremented by one sample and the learning process is repeated on the expanded training set. The structural size, structural life time and hop size (see definitions in section III B) are measured and averaged over many online learning histories.

Table 1 quantifies the evolution of structural size and life time for a progression of training set sizes. Each row of the table presents the stability indicators for a given range of training set sizes averaged over 1000 online training histories with data drawn from the ground truth distribution. We assume that all simulated observations are accurate so that there is no uncertainty in the data. For this experiment we used a model with 12 redundant variables.

| Training set range | Average structural size | Average structural life time |
|---|---|---|
| 0 - 19 | 3,69 ± 1,83 | 3,2 ± 2,08 |
| 20 -39 | 7,24 ± 1,06 | 7,52 ± 5,01 |
| 40 - 59 | 8,79 ± 0,89 | 12,84 ± 8,31 |
| 60 - 79 | 9,73 ± 0,85 | 18,94 ± 11,70 |
| 80 - 99 | 10,38 ± 0,80 | 26,72 ± 15,84 |
| 100 -119 | 10,86 ± 0,73 | 34,82 ± 19,62 |
| 120 -139 | 11,17 ± 0,64 | 41,52 ± 22,44 |
| 140 - 159 | 11,44 ± 0,58 | 53,42 ± 27,96 |
| 160 179 | 11,64 ± 0,50 | 63,38 ± 28,28 |
| 180 -199 | 11,84 ± 0,37 | 84,18 ± 37,81 |
| 200 - 219 | 11,87 ± 0,34 | 112 ± 29,80 |
| 220 - 239 | 11,90 ± 0,30 | 126 ± 10,65 |

*Table 1. Average structural size and structural life time for the preferred solution according to the learning algorithm described in section II A. The standard deviation d of the tabulated values is indicated by (± d). The data is batched over a range of training set sizes in order to make trends visible in spite of the large variance of the structural life time.*

Note that only false predictors contribute to the data shown in Table 1. The learning process is terminated as soon as the system finds the ground truth hypothesis. Equation (9) gives an idea of how many training samples that are be needed to suppress the false predictors so that the algorithm captures the relevant solution. In the experiments we found that the average number of training samples needed for exiting the false predictor phase is $116 ± 61$ where again the standard deviation is indicated. In the experiments we also measured the hop size and found that the average hop size always is very close to unity independently of the training set size.

Our experiments illustrate and confirm the effects that was presaged in section III namely that both the size and the life time of redundant structures increases with increasing training size in the false predictor phase.

## V. CONCLUSIONS

Comparing the theoretical and experimental results on the toy model with the rule of thumb suggested in section II C we note that the evolution of solution structure in the zone of false predictors is dangerously close to satisfying all the suggested stability requirements. False predictors can masquerade as relevant solutions since,

*1) The life time of redundant structures increases with the number of redundant variables and might be longer than the maintenance interval. In a real-life situation it is impossible to assess if this is the case since the true joint probability distribution is unknown and the relational structure is much more complex than in the toy model.*
*2) Redundant structures suffer only minor refinements for the later part of the false predictor phase.*
*3) The life time of consecutive redundant structures increases as training examples accumulate.*

Redundant structures imitate our expectations for valid solutions mainly because larger structures have an exponentially larger number of associated CPTs that take a progressivly longer time to falsify. Obviously it is only in the case of binary CPTs that we can talk about the number of CPTs but the survivablity of redundant structures is actually expected to increase for real-valued CPTs since the parameter space is less constrained. The masquerading effect is hence expected to preavail for generic SRL models and the effect is in fact unique for relational models. More specialized redundant hypotheses are in general easier to falsify with new data in classical machine learning while more specialized redundant SRL structures are harder to falsify.

The BLP learning algorithm that is briefly reviewed in section II A and comprehensibly described in [11] could be amended by more sophisticated search and regularization methods although the scope for such improvements is severely limited by computational efficiency issues. Replacing greedy hill climbing with search techniques designed for a rough fitness landscape, such as simulated annealing, would reduce the risk of ending up in a false optimum. Note, however, that false predictors typically are global optima in the false predictor phase and hence prevail even if robust search techniques are applied. The learning algorithm of the KReator toolkit [16] includes simple regularization since the default setting limits the number of predicates in the body of a Bayesian clause to three. A wide range of alternative methods for discriminating against too complex solutions can be tried including replacing the hard cut-off with a soft weighting and also entropy-based methods. It is, however, often difficult to estimate the complexity of the wanted solution. Too strict regularization could emasculate the expressive power of SRL.

No foreseeable modification of the learning algorithm is hence a panacea for the online stability issue. Better search methods are not a certain remedy since false predictors can have a lower loss function value than valid solutions. Regularization favors simple solutions but the main argument for SRL is the ability to model complex real-world relational structures so simplicity cannot be driven too far. Presently it appears that online learning stability can be guaranteed only by selecting non-relational machine learning methods with convex loss functions where the form of the convergence rate (6) is known. Relational algorithms such as BLP are not provable online stable and our analysis indicates that false predictor structures can behave as bona fide solutions according to our rules of thumb in section II C.

The online learning stability of SRL algorithms is a therefore in doubt since real-life SRL scenarios often include a large number of variables, a limited supply of training examples and little a priori knowledge about the ground truth probability distribution. Since the underlying probability distribution is unknown it is difficult to estimate how much initial training data that is needed to eradicate false predictors and generate a relevant hypothesis. Because the life time of false predictors is unknown it is difficult to know when the learning process has converged to a hypothesis that can be trusted to be sufficiently stable according to the operational requirements that were discussed in section I E. These problems remain even in an ideal case where the algorithm fortuitously is supplied with the correct model for example from human expert. As the algorithm continues to learn online it can still get lost in the jagged fitness landscape

Should one in spite of these issues select to use a SRL-based solution in e.g. a surveillance application we can presently just offer the following advice for enhancing online stability.

1) Design a loss function that reflects operational requirements (rather than some default loss function of the SRL package).

2) Approve the system for operational use only when all of the following conditions are satisfied,

   a) The rules of thumb in section II C are satisfied.

   b) The solution gives satisfactory performance from an operational point of view.

   c) The selected structure is consistent with an appropriate domain knowledge base (see discussion in section I C)

   d) The selected structure makes sense according to human expert evaluation that is repeated for each maintenance interval.

The first requirement ensures that online training is directed to improving operational performance while the second requirement reduces the risk for launching a live system while the learning process still is thrashing in the false predictor phase.

Although it appears that human monitoring may help us to establish at least some partial control of online stability practical applications often include other pressing stability issues. In this paper we have tacitly assumed that there is a stable underlying joint probability distribution that is explored by a growing data set. Many applications include probability distributions that change over multiple and initially unknown time scales. The very concept of stochastic modeling is also challenged in scenarios including intelligent adversarial opposition. It appears that there are good reasons for further investigating the stability of SRL online learning under conditions that are relevant for real-life applications.

ACKNOWLEDGMENT

Helpful advice on BLP learning and on the KReator toolkit from Matthias Thimm is gratefully acknowledged. This research was supported by the SecurityLink strategic research centre as well as by the European Commission under Grant No 261679 (CONTAIN).